# LG-LSQ: Learned Gradient Linear Symmetric Quantization

Shih-Ting Lin[1]*, Zhaofang Li[1]*, Yu-Hsiang Cheng[1], Hao-Wen Kuo[1], Chih-Cheng Lu[2],
Kea-Tiong Tang[1]
[1]Department of Electrical Engineering, National Tsing Hua University
[2]Information and Communication Labs Industrial Technology Research Institute
kttang@ee.nthu.edu.tw
(*These authors contributed equally to this work)

## Abstract

Deep neural networks with lower precision weights and operations at inference time have advantages in terms of the cost of memory space and accelerator power. The main challenge associated with the quantization algorithm is maintaining accuracy at low bit-widths. We propose learned gradient linear symmetric quantization (LG-LSQ) as a method for quantizing weights and activation functions to low bit-widths with high accuracy in integer neural network processors. First, we introduce the scaling simulated gradient (SSG) method for determining the appropriate gradient for the scaling factor of the linear quantizer during the training process. Second, we introduce the arctangent soft round (ASR) method, which differs from the straight-through estimator (STE) method in its ability to prevent the gradient from becoming zero, thereby solving the discrete problem caused by the rounding process. Finally, to bridge the gap between full-precision and low-bit quantization networks, we propose the minimize discretization error (MDE) method to determine an accurate gradient in backpropagation. The ASR+MDE method is a simple alternative to the STE method and is practical for use in different uniform quantization methods. In our evaluation, the proposed quantizer achieved full-precision baseline accuracy in various 3-bit networks, including ResNet18, ResNet34, and ResNet50, and an accuracy drop of less than 1% in the quantization of 4-bit weights and 4-bit activations in lightweight models such as MobileNetV2 and ShuffleNetV2.

## 1 Introduction

Deep learning technologies have exhibited excellent performance in various fields, such as computer vision [Krizhevsky, 2012] and natural language processing [Luong et al., 2015]. However, applying deep learning technologies in such fields requires enormous memory storage and computational power, thereby increasing the power consumption of the hardware used. Model compression methods, such as quantization, can be used to increase the operational efficiency of the deep neural networks (DNNs) on the hardware and reduce the computational burden on the hardware [Gholami et al., 2021].

A high-quality network quantizer enables neural network processors to manage low-precision integer operations and minimize accuracy loss. DoReFa-Net [Zhou et al., 2016] is a quantization method designed for low-bit hardware accelerators that can be used to quantize networks of any number of bit-width and deploy the quantized networks to CPUs, GPUs, application-specific integrated circuits (ASICs), and field-programmable gate arrays (FPGAs). However, this method does not account for batch normalization [Ioffe and Szegedy, 2015]; therefore, new architectures cannot use it for network compression. Progressive-freezing iterative training (PROFIT) [Park and Yoo, 2020] uses differentiable and unified quantization (DuQ), which evaluates the effects of different network layers on accuracy, to compress a MobileNet DNN [Sandler et al., 2018] to 4 bits. Layers with more negligible effects on accuracy to that layer receive more training. Although learned linear symmetric quantization (LLSQ) [Zhao et al., 2019] can be used to make a network wholly deployable on ASIC hardware, the search method of the scaling factor in LLSQ remains suboptimal.

While quantizing the full-precision value to a low-bit value, rounding functions cause accuracy loss due to discontinuous distribution. Because the rounding process is nondifferentiable, the straight-through estimator (STE) [Bengio et al., 2013] is often used to propagate the same gradient without considering discretization errors between inputs and outputs. In such cases, the slope is not representative of the actual situation, thereby resulting in loss of accuracy [McKinstry et al., 2018; Li et al., 2017].

To solve the aforementioned problems and minimize quantization errors, we propose the learned gradient linear symmetric quantization (LG-LSQ) method. First, we introduce the scaling simulated gradient (SSG) method to optimize the adjustment of the gradient scaling factor, thereby enabling the network to identify more suitable quantization parameters during the training process. Second, we propose using the arctangent soft round (ASR) method instead of the STE or rounding function to maintain the differentiability while still enabling the backpropagation of the gradient through differentiation. Finally, we introduce the minimize discretization error (MDE) method in the training process to reduce the

quantization errors at low bit-widths. The main contributions of our work are as follows:
1. The LG-LSQ method can be used to quantize both activations and weights and outperforms the state-of-the-art quantization methods.
2. Quantization errors can be minimized by using the SSG method to train the learnable scaling factor and using the ASR and MDE methods instead of STE quantization methods.
3. The proposed method is more effective than other uniform quantization methods, such as DoReFa-Net and LLSQ. We tested our method by using various neural network architectures with both CIFAR-10 and ImageNet datasets and determined that it exhibits significantly higher accuracy than do the aforementioned quantization methods.

## 2 Related Works

### 2.1 Network Quantization

Quantization methods can be categorized as linear or nonlinear. Although many nonlinear quantization methods exhibit high performance at low bit-widths, they require additional conversions to obtain the correct quantization results. For example, Han *et al*. [2016] and Park *et al*., [2017] used the lookup table method to transform the quantized values. To avoid unnecessary additional transformations, linear quantization is essential for state-of-the-art accelerators.

Linear quantization can be categorized as symmetric or asymmetric. The calculations involved in asymmetric quantization are more complicated than those involved in symmetric quantization. For example, asymmetric quantization requires additional addition and subtraction operations and linear operations before the multiplication and requires one more parameter (zero-point) than does symmetric quantization [Krishnamoorthi, 2018]. Because they do not require the design of additional functions, linear symmetric quantization methods are optimal for implementation on hardware accelerators.

### 2.2 STE

Standard quantization methods require the use of two mathematical models on DNNs: a numerical reducer and an STE. The most common examples of numerical reducers are sign functions and rounding functions, which are used in forward propagation to quantize the input signal to its corresponding value. The STE is applied in backpropagation. However, some zero and nondifferentiable values result from employing the rounding function in quantization. To enable the quantization network to calculate and transfer the gradient, the STE defines the value of the rounding function in backpropagation, and the following is its mathematical formula:

$$\frac{\partial L}{\partial x} = \frac{\partial L}{\partial x^q} \qquad (1)$$

where $L$ is the loss function, $x$ is the original value, and $x^q$ is the quantized value. Because the STE directly maps the gradient identity of the quantization function to the actual value, the network gradient of the deep neural network can be effectively transmitted back. However, this approach cannot account for quantization errors [McKinstry *et al*., 2018; Li *et al*., 2017] because the derivative value is the same regardless of how many bits are quantized. To solve the aforementioned problems, QuantNoise [Fan *et al*., 2020] uses unbiased gradients to quantize a random subset of weights. However, a major disadvantage of stochastic quantization methods is the overhead of generating random numbers for every weight update; therefore, such methods have not yet been widely adopted. Differentiable soft quantization (DSQ) [Gong *et al*., 2019] uses the tanh function to simulate the rounding process, whereas relaxed quantization (RQ) [Louizos *et al*., 2019] uses a special numerical reducer to replace the STE. However, the hyperparameters must be specially adjusted; otherwise, gradient disappearance or explosion becomes likely.

## 3 Proposed Methods

Section 3.1 introduces our quantization method, LG-LSQ, for weights and activation functions. Section 3.2 explains how to use the SSG method to determine how to adjust the gradient of the scaling factor. Section 3.3 introduces how the ASR method can be used as an alternative to the rounding function and STE. Finally, section 3.4 explains how the MDE can be used to further improve the process described in section 3.3 to maintain high accuracy when quantizing to low bit-widths.

### 3.1 Preliminary

We propose a LG-LSQ method suitable for weights and activation functions. Channel-wise and layer-wise quantization are used in the convolutional and fully-connected layers, respectively. $\mathbf{quantize}_k(x_i^r, \boldsymbol{\alpha})$ quantizes a real number input $x_i^r$ to a $k$-bit number output $x_i^q$ and can be defined as follows:

$$x_i^q = \frac{x_i^t}{\alpha} = clamp\left(round\left(\frac{x_i^r}{\alpha}\right), Q_{min}, Q_{max}\right) \qquad (2)$$

where $x_i^r \in R$ is the input value that has not yet been quantized in the weight or activation function of the $i$th layer; $\alpha \in R^+$ is the scaling factor, known as the quantizer step size; $Q_{min}$ and $Q_{max}$ represent the number of positive and negative quantization levels, respectively; and the clamp($s$, $Q_{min}, Q_{max}$) and round($\cdot$) functions are the clipping function and rounding function of the numerical reducer, respectively. The clamp($s$, $Q_{min}, Q_{max}$) function returns $s$, with values below $Q_{min}$ set to $Q_{min}$ and values above $Q_{max}$ set to $Q_{max}$. If the activation function is a ReLU function, the activations are nonnegative values; therefore, we clamp the values to $[0, 2^{bit} - 1]$ to obtain $x_i^t \in \{0, \alpha, \cdots, (2^{bit-1} - 1)\alpha\}$ and $x_i^q \in \{0, 1, \cdots, (2^{bit-1} - 1)\}$, respectively.

## 3.2 SSG

In the LLSQ training method [Zhao *et al.*, 2019], the gradient of the quantization parameter is searched with a fixed value, and the optimal value may therefore not be obtained. To address this problem, we use the SSG to adjust the gradient of the quantization parameter to enahnce the flexibility of the scaling factor for self-learning. This method is defined as follows:

$$g_\alpha = -\alpha^2 \cdot (\arg\,min([E_l, E_m, E_r]) - 1) \quad (3)$$

where the gradient of the scaling factor is represented by $g_\alpha \in \{-\alpha^2, 0, \alpha^2\}$; $\arg min([E_l, E_m, E_r]) \in \{0, 1, 2\}$; and $E_l, E_m,$ and $E_r$ are calculated as follows:

$$E_l = \sum_i (x_i^r - quantize_k(x_i^r, \alpha z_l) \cdot \alpha z_l)^2$$
$$E_m = \sum_i (x_i^r - quantize_k(x_i^r, \alpha) \cdot \alpha)^2$$
$$E_r = \sum_i (x_i^r - quantize_k(x_i^r, 2\alpha z_r) \cdot 2\alpha z_r)^2$$
$$z_l = 0.5 + z, z_r = 1 - z \quad (4)$$

where $z_l$ and $z_r$ are the parameters that affect the gradient of the quantization parameter, and $z_l$ and $z_r$ are determined by the magnitude of each update $z$. Under the initial conditions, $z = 0$ and $[0.5\alpha, \alpha, 2\alpha]$ are used to determine the value of $[E_l, E_m, E_r]$. As the number of training iterations increases, the value of $\arg min([E_l, E_m, E_r])$ is set to 0 or 2 a certain number of consecutive times, the initially selected $0.5\alpha$ or $2\alpha$ must be updated toward $\alpha$. Therefore, SSG is a flexible method that allows for determination of the optimal quantization parameter.

## 3.3 ASR

Rounding functions, which are often used in the quantization process, are nondifferentiable. Therefore, in training quantized DNN, the STE method is often used to handle the problem of non-differentiability caused by the rounding function. However, this approach increases the quantization error because of the use of approximate operations to equate the input and output gradients. To solve these problems, we propose using the ASR method, for which the forward propagation method is as follows:

$$x_{ASR} = x_{floor} + \frac{1}{\pi}(\tan^{-1}(\lambda(x - x_{floor} - \frac{1}{2})) + \frac{\pi}{2}) \quad (5)$$

where $x_{floor} = floor(x)$; floor($\cdot$) is the floor function; $\lambda$ is a hyperparameter used to determine the slope of the ASR; and $tan^{-1}(\cdot)$ is a logical function in which larger input values yield output values closer to $0.5\pi$, smaller input values yield output values closer to $-0.5\pi$, and an input value of 0 yields an output value of 0. Because $x \geq x_{floor}$, we obtain $(x - x_{floor}) \in [0,1]$. We subtract a bias value of 0.5 to obtain $(x - x_{floor} - 0.5) \in [-0.5, +0.5]$; subsequently, by adjusting $\lambda$, we obtain $tan^{-1}\big(\lambda \cdot (x - x_{floor} - 0.5)\big) \in [-0.5\pi, 0.5\pi]$. By shifting and scaling the value, we can obtain the formula $1/\pi \cdot \big(tan^{-1}\big(\lambda \cdot (x - x_{floor} - 0.5)\big) + 0.5\pi\big) \in [0,1]$. Finally, we add $x_{floor}$ to obtain the final output value in a process similar to rounding. As $\lambda$ increases, the ASR function becomes closer to the original rounding function.

The backward propagation function is as follows:

$$g_{x_{ASR}} = \frac{\partial x_{ASR}}{\partial x} = \frac{\lambda}{\pi(1+m^2)}, m = \lambda\left(x - x_{floor} - \frac{1}{2}\right) \quad (6)$$

where $m = \lambda(x - x_{floor} - 0.5)$ and $x_{floor}$ is the floor($\cdot$) function. When $\lambda = 1$, we can find the distribution of the differential value of the ASR function because when the input value is nearer an integer, the smaller the differential value is, the smaller the step size for the input value becomes. Because this method allows for differentiation, it bridges the gap between a full-precision value and low-bit value.

## 3.4 MDE

To reduce the quantization error rate, we reference the element-wise gradient scaling (EWGS) technique [Lee *et al.*, 2021] and propose the MDE method, which adds a penalty term to the original gradient $g_{x_{ASR}}$ and for which the formula is as follows:

$$g_x = g_{x_{ASR}} + tanh(g_{x_{ASR}}) \cdot x_{ERROR} \cdot g_{x_{ASR}}$$
$$= g_{x_{ASR}} \cdot (1 + tanh(g_{x_{ASR}}) \cdot x_{ERROR}) \quad (7)$$
$$x_{ERROR} = x - x_{ASR}$$

where $g_{x_{ASR}}$ is the gradient of ASR; $x_{ERROR} = x - x_{ASR}$ is the error term before and after rounding; and $tanh(\cdot)$ is a logic function in which larger input values yield output values closer to 1, smaller input values yield output values closer to $-1$, and an input value of 0 yields an output value of 0. Therefore, considering the effect of the positive and negative values produced by the gradient of the ASR, we use the characteristics of $tanh(g_{x_{ASR}}) \in [-1,1]$ to reduce the quantization error rate and the differences between the full-precision and quantized values.

## 4 Experimental Results

This section presents the results divided into four parts: SSG, ASR, MDE, and LG-LSQ (that is, our integration of the SSG, ASR, and MDE methods). The methods used in sections 4.1, 4.2, and 4.3 involve linear symmetric quantization of weights and activation functions, in which the first and last layers maintain full precision. In section 4.4, we also quantize the entire network, including the first and last layers. When the CIFAR-10 dataset is used, we train the network from scratch for 300 epochs and set the initial learning rate value to 2e-2. When the ImageNet [Russakovsky *et al.*, 2015] dataset is used, we use pretrained weights; train the network for 90 epochs using a warmup learning scheduler in the first three

| Model | W/A | Simulated Gradient | | Scaling Simulated Gradient | |
|---|---|---|---|---|---|
| | | Top-1(%) | Top-5(%) | Top-1(%) | Top-5(%) |
| ResNet18 | 32/32 | 69.76 | 89.08 | - | - |
| | 4/4 | 69.84 | 89.14 | **70.02** | **89.16** |
| | 3/3 | 68.08 | 88.20 | **68.44** | **88.35** |
| Mobile-NetV2 | 32/32 | 71.80 | 90.37 | - | - |
| | 6/6 | 71.20 | 89.99 | **71.31** | **90.12** |
| | 5/5 | 70.45 | 89.69 | **70.54** | 89.64 |
| | 4/4 | 67.37 | 87.99 | **68.59** | **88.46** |
| Shuffle-NetV2 | 32/32 | 69.36 | 88.32 | - | - |
| | 8/8 | 68.46 | 87.79 | **68.75** | **88.15** |
| | 4/4 | 61.86 | 83.40 | **62.23** | **83.73** |

Table 1: Comparison of simulated gradient methods on ImageNet. Top-1 and Top-5 accuracy (%) are provided.

| | | DoReFa-Net | |
|---|---|---|---|
| Model | W/A | Round + STE | ASR |
| | | Accuracy (%) | Accuracy (%) |
| VGG7 (Ref: 93.52) | 8/8 | 93.33 | **93.68** |
| | 8/4 | 93.28 | **93.32** |
| | 4/4 | 93.21 | **93.25** |
| VGG16 (Ref: 93.84) | 8/8 | 93.72 | **93.83** |
| | 8/4 | 93.61 | **93.79** |
| | 4/4 | **93.56** | 93.51 |
| ResNet20 (Ref: 92.71) | 8/8 | 92.46 | **92.57** |
| | 8/4 | 92.24 | **92.28** |
| | 4/4 | **92.03** | 92.02 |
| ResNet18 (Ref: 93.01) | 8/8 | 92.90 | **92.98** |
| | 8/4 | 92.87 | **92.93** |
| | 4/4 | 92.84 | **92.88** |
| Mobile-NetV2 (Ref: 94.46) | 8/8 | 94.31 | **94.50** |
| | 8/4 | 94.24 | **94.41** |
| | 4/4 | 94.00 | **94.37** |

Table 2: Comparison of ASR and Round+STE methods on CIFAR-10 in DoReFa-Net quantization.

epochs and a cosine scheduler in the remaining epochs; and set the learning rate value to 5e-5.

### 4.1 SSG

The SSG method can update the searching range of simulated gradients under periodic checks. Therefore, we set the checking iteration period $iter_{target}$ to one-fifth of the iterations per epoch during the training process. For example, when the ImageNet dataset is being used, the batch size is 256, and approximately 5000 iterations are completed per epoch during the training process; therefore, we set $iter_{target}$ to 1000. Whenever the number of iterations reaches $iter_{target}$ during training, the continual repetitions of $\arg min([E_l, E_m, E_r])$ are calculated. Once $E_l$ or $E_r$ appears four times consecutively, the original value of $0.5\alpha$ or $2\alpha$ must be updated by the amplitude $z$ toward $\alpha$. In our experiment, $z = 0.03125$; that is, the amplitude of each update is $z$. A limit is used to ensure that the value of $z_l$ is not greater than $0.5\alpha$ and that the value of $z_r$ is not less than $0.5\alpha$ in the training process.

| | | LLSQ | |
|---|---|---|---|
| Model | W/A | Round + STE | ASR |
| | | Accuracy (%) | Accuracy (%) |
| VGG7 (Ref: 93.59) | 8/8 | 93.53 | **93.64** |
| | 8/4 | 93.32 | **93.51** |
| | 4/4 | 93.13 | **93.31** |
| VGG16 (Ref: 94.04) | 8/8 | 93.76 | **93.97** |
| | 8/4 | 93.62 | **93.87** |
| | 4/4 | **93.57** | 93.52 |
| ResNet20 (Ref: 92.74) | 8/8 | 92.66 | **92.68** |
| | 8/4 | 92.26 | **92.54** |
| | 4/4 | 92.15 | **92.52** |
| ResNet18 (Ref: 93.07) | 8/8 | 93.05 | **93.15** |
| | 8/4 | 92.93 | **93.06** |
| | 4/4 | 92.18 | **93.00** |
| MobileNetV2 (Ref: 94.78) | 8/8 | 94.36 | **94.86** |
| | 8/4 | 94.29 | **94.64** |
| | 4/4 | 94.19 | **94.43** |

Table 3: Comparison of ASR and Round+STE methods on CIFAR-10 in LLSQ quantization.

| | | DoReFa-Net | |
|---|---|---|---|
| Model | W/A | ASR | ASR + MDE |
| | | Accuracy (%) | Accuracy (%) |
| VGG7 (Ref: 93.52) | 8/8 | 93.68 | **93.75** |
| | 8/4 | 93.32 | **93.43** |
| | 4/4 | 93.25 | **93.42** |
| VGG16 (Ref: 93.84) | 8/8 | 93.83 | **94.04** |
| | 8/4 | 93.79 | **93.81** |
| | 4/4 | 93.51 | **93.63** |
| ResNet20 (Ref: 92.71) | 8/8 | 92.57 | **92.74** |
| | 8/4 | 92.28 | **92.65** |
| | 4/4 | 92.02 | **92.60** |
| ResNet18 (Ref: 93.01) | 8/8 | 92.98 | **93.21** |
| | 8/4 | 92.93 | **93.07** |
| | 4/4 | 92.88 | **93.01** |
| MobileNetV2 (Ref: 94.46) | 8/8 | 94.50 | **94.56** |
| | 8/4 | 94.41 | **94.53** |
| | 4/4 | 94.37 | **94.44** |

Table 4: Comparison of ASR and ASR+MDE methods on CIFAR-10 in DoReFa-Net quantization.

Using ImageNet [Russakovsky *et al.*, 2015], we tested various network architectures, namely ResNet18 [He *et al.*, 2016], MobileNetV2 [Sandler *et al.*, 2018], and ShuffleNetV2 [Ma *et al.*, 2018]. As indicated by the data presented in Table 1, our approach yielded more accurate results

| | LLSQ | | |
|---|---|---|---|
| Model | W/A | ASR | ASR + MDE |
| | | Accuracy (%) | Accuracy (%) |
| VGG7 (Ref: 93.59) | 8/8 | 93.64 | **93.83** |
| | 8/4 | 93.51 | **93.70** |
| | 4/4 | 93.31 | **93.56** |
| VGG16 (Ref: 94.04) | 8/8 | 93.97 | **94.01** |
| | 8/4 | 93.87 | **93.91** |
| | 4/4 | 93.52 | **93.58** |
| ResNet20 (Ref: 92.74) | 8/8 | 92.68 | **92.72** |
| | 8/4 | 92.54 | **92.63** |
| | 4/4 | 92.52 | **92.55** |
| ResNet18 (Ref: 93.07) | 8/8 | 93.15 | **93.17** |
| | 8/4 | 93.06 | **93.10** |
| | 4/4 | 93.00 | **93.01** |
| MobileNetV2 (Ref: 94.78) | 8/8 | 94.86 | **95.02** |
| | 8/4 | 94.64 | **94.97** |
| | 4/4 | 94.43 | **94.73** |

Table 5: Comparison of ASR and ASR+MDE methods on CIFAR-10 in LLSQ quantization.

| | | LLSQ | | | |
|---|---|---|---|---|---|
| Model | W/A | Round + STE | | ASR + MDE | |
| | | Top-1(%) | Top-5(%) | Top-1(%) | Top-5(%) |
| ResNet18 | 32/32 | 69.76 | 89.08 | - | - |
| | 4/4 | 69.84 | 89.14 | **70.57** | **89.54** |
| | 3/3 | 68.08 | 88.20 | **69.96** | **89.28** |
| ResNet34 | 32/32 | 73.30 | 91.42 | - | - |
| | 4/4 | 73.60 | 91.28 | **73.72** | **91.55** |
| MobileNetV2 | 32/32 | 71.80 | 90.37 | - | - |
| | 6/6 | 71.20 | 89.99 | **71.49** | **90.22** |
| | 5/5 | 70.45 | 89.69 | **71.45** | **90.17** |
| | 4/4 | 67.37 | 87.99 | **71.26** | **89.98** |
| ShuffleNetV2 | 32/32 | 69.36 | 88.32 | - | - |
| | 8/8 | 68.46 | 87.79 | **68.59** | **88.22** |
| | 4/4 | 61.86 | 83.40 | **67.96** | **87.62** |

Table 6: Comparison of Round+STE and ASR+MDE methods on ImageNet in LLSQ quantization. Top-1 and Top-5 accuracy (%) are given.

than did the simulated gradient method used in LLSQ [Zhao et al., 2019].

### 4.2 ASR

The ASR method can be applied to the rounding function in various uniform quantization approaches. $\lambda$ is assigned an initial value, and as the number of training epochs increases, the value of $\lambda$ gradually increases. A limit is used to prevent the value of $\lambda$ from becoming too large and causing overfitting. As indicated by the data in Tables 2 and 3, regardless of whether the DoReFa-Net [Zhou et al., 2016] or LLSQ [Zhao

| Architecture | Methods | W/A | Top-1 | Top-5 |
|---|---|---|---|---|
| | Full-precision | 32/32 | 71.80 | 90.37 |
| | LLSQ | 6/6 | 71.20 | 89.99 |
| | LLSQ | 5/5 | 70.45 | 89.69 |
| | DSQ | 4/4 | 64.80 | - |
| MobileNetV2 | LLSQ | 4/4 | 67.37 | 87.99 |
| | EWGS | 4/4 | 70.30 | - |
| | **LG-LSQ(Ours)** | **6/6** | **71.68** | **90.22** |
| | **LG-LSQ(Ours)** | **5/5** | **71.73** | **90.20** |
| | **LG-LSQ(Ours)** | **4/4** | **71.58** | **90.18** |
| | **LG-LSQ(Ours)** | **3/3** | **71.31** | **90.01** |
| | FAT | 5/5 | 69.60 | 89.20 |
| | FAT | 4/4 | 69.20 | 88.90 |
| | LCQ* | 4/4 | 70.80 | 89.70 |
| MobileNetV2** | FAT | 3/3 | 62.80 | 84.90 |
| | **LG-LSQ(Ours)** | **6/6** | **71.32** | **90.05** |
| | **LG-LSQ(Ours)** | **5/5** | **71.25** | **89.89** |
| | **LG-LSQ(Ours)** | **4/4** | **71.09** | **89.79** |
| | **LG-LSQ(Ours)** | **3/3** | **70.62** | **89.72** |
| | Full-precision | 32/32 | 69.36 | 88.32 |
| | LLSQ | 8/8 | 68.46 | 87.79 |
| ShuffleNetV2 | LLSQ | 4/4 | 61.86 | 83.4 |
| | **LG-LSQ(Ours)** | **8/8** | **69.07** | **88.27** |
| | **LG-LSQ(Ours)** | **4/4** | **68.31** | **87.93** |
| | **LG-LSQ(Ours)** | **3/3** | **67.38** | **87.26** |
| ShuffleNetV2** | **LG-LSQ(Ours)** | **8/8** | **69.03** | **88.24** |
| | **LG-LSQ(Ours)** | **4/4** | **67.88** | **87.53** |

\* First and last layers with 8 bits.
\*\* First and last layers are quantized.
Table 7: Comparison of state-of-the-art quantization methods on ImageNet in lightweight models.

et al., 2019] quantization method was employed, ASR exhibited performance superior to that of the traditional rounding function and STE method in the quantization of 8-bit weights and 8-bit activations, 8-bit weights and 4-bit activations, and by most measures, of 4-bit weights and 4-bit activations. In a few models, because the number of quantized bits is low, overfitting is prone to occur when the hyperparameters are too large. The MDE method presented in section 3.4 serves as a potential solution to this problem.

### 4.3 MDE

The MDE method can be applied in various approaches to linear symmetric quantization, and the parameter settings are the same as those presented in section 4.2. As indicated in Tables 4–6, the ASR plus MDE (ASR+MDE) method exhibited performance superior to that of the ASR method for various neural networks, regardless of whether the CIFAR-10 or ImageNet [Russakovsky et al., 2015] dataset was used.

### 4.4 LG-LSQ (SSG+ASR+MDE)

We compared LG-LSQ with state-of-the-art approaches, namely RQ [Louizos et al., 2019], UNIQ [Baskin et al., 2018], PACT [Choi et al., 2018], LQ-Nets [Zhang et al., 2018], DSQ [Gong et al., 2019], LLSQ [Zhao et al., 2019],

| Architecture | Methods | W/A | Top-1 | Top-5 |
|---|---|---|---|---|
| ResNet18 | Full-precision | 32/32 | 69.76 | 89.08 |
| | PACT | 5/5 | 69.80 | 89.30 |
| | LLSQ | 4/4 | 69.84 | 89.14 |
| | QIL | 4/4 | 70.10 | - |
| | LLSQ | 3/3 | 68.08 | 88.20 |
| | LQ-Nets | 3/3 | 68.20 | 87.90 |
| | DSQ | 3/3 | 68.66 | - |
| | QIL | 3/3 | 69.20 | - |
| | EWGS | 3/3 | 69.70 | - |
| | **LG-LSQ(Ours)** | **4/4** | **70.78** | **89.77** |
| | **LG-LSQ(Ours)** | **3/3** | **70.31** | **89.55** |
| ResNet18** | RQ | 8/8 | 69.97 | 89.44 |
| | UNIQ | 5/8 | 68.00 | - |
| | DSQ | 4/4 | 69.56 | - |
| | FAQ* | 4/4 | 69.78 | 89.11 |
| | LLSQ | 3/3 | 66.67 | 87.42 |
| | FAT | 3/3 | 69.00 | 88.60 |
| | LSQ+ | 3/3 | 69.40 | - |
| | **LG-LSQ(Ours)** | **4/4** | **70.24** | **89.38** |
| | **LG-LSQ(Ours)** | **3/3** | **69.47** | **88.90** |
| ResNet34 | Full-precision | 32/32 | 73.30 | 91.42 |
| | DSQ | 4/4 | 72.76 | - |
| | LLSQ | 4/4 | 73.60 | 91.28 |
| | QIL | 4/4 | 73.70 | - |
| | LLSQ | 3/3 | 72.02 | 90.66 |
| | DSQ | 3/3 | 72.54 | - |
| | QIL | 3/3 | 73.10 | - |
| | EWGS | 3/3 | 73.30 | - |
| | **LG-LSQ(Ours)** | **4/4** | **73.92** | **91.65** |
| | **LG-LSQ(Ours)** | **3/3** | **73.56** | **91.43** |
| ResNet34** | LLSQ | 4/4 | 72.94 | 91.20 |
| | FAQ* | 4/4 | 73.31 | 91.32 |
| | LLSQ | 3/3 | 70.97 | 89.95 |
| | FAT | 3/3 | 73.20 | 91.20 |
| | LSQ* | 3/3 | 73.40 | 91.40 |
| | **LG-LSQ(Ours)** | **4/4** | **73.75** | **91.63** |
| | **LG-LSQ(Ours)** | **3/3** | **73.48** | **91.40** |
| ResNet50 | Full-precision | 32/32 | 76.15 | 92.87 |
| | LQ-Nets | 4/4 | 75.10 | 92.40 |
| | PACT | 4/4 | 76.50 | 93.30 |
| | LQ-Nets | 3/3 | 74.20 | 91.60 |
| | **LG-LSQ(Ours)** | **4/4** | **76.75** | **93.20** |
| | **LG-LSQ(Ours)** | **3/3** | **76.47** | **93.07** |
| ResNet50** | LCQ* | 4/4 | 76.60 | - |
| | LSQ* | 4/4 | 76.70 | 93.20 |
| | LSQ* | 3/3 | 75.80 | 92.70 |
| | LCQ* | 3/3 | 76.30 | - |
| | **LG-LSQ(Ours)** | **4/4** | **76.73** | **93.19** |
| | **LG-LSQ(Ours)** | **3/3** | **76.41** | **93.10** |

\* First and last layers with 8 bits.
\*\* First and last layers are quantized.
Table 8: Comparison of state-of-the-art quantization methods on ImageNet in various networks.

QIL [Jung *et al.*, 2019], FAT [Tao *et al.*, 2021], APoT [Li *et al.*, 2020], LSQ+ [Bhalgat *et al.*, 2020], EWGS [Lee *et al.*, 2021], LSQ [Esser *et al.*, 2020], PROFIT [Park and Yoo, 2020], and LCQ [Yamamoto, 2021]. The proposed quantizer achieved full-precision baseline accuracy with various 3-bit networks, including ResNet18, ResNet34, and ResNet50 and achieved an accuracy drop of less than 1% for 4-bit weights and 4-bit activations with lightweight models, such as MobileNetV2 and ShuffleNetV2. Tables 7 and 8 present the detailed results of the comparison.

## 5 Conclusions

In this work, we proposed LG-LSQ as a network quantization algorithm to bridge the gap between quantized and full-precision networks. The SSG method can determine the optimal scaling parameters during network training through linear symmetric quantization. The ASR method can use differentiable features to preserve gradient features during training. Finally, the MDE method can reduce the quantization error rate during the training process. Our method can be easily integrated with diverse CNN architectures and linear symmetric quantization methods. The results indicate that LG-LSQ outperforms conventional state-of-the-art methods and narrows the gap between low-bit and full-precision models for image classification.